\documentclass[]{assets/latex/ieee}
\usepackage{lmodern}
\usepackage{amssymb,amsmath}
\usepackage{ifxetex,ifluatex}
\usepackage{fixltx2e} 
\ifnum 0\ifxetex 1\fi\ifluatex 1\fi=0 
  \usepackage[T1]{fontenc}
  \usepackage[utf8]{inputenc}
\else 
  \ifxetex
    \usepackage{mathspec}
    \usepackage{xltxtra,xunicode}
  \else
    \usepackage{fontspec}
  \fi
  \defaultfontfeatures{Mapping=tex-text,Scale=MatchLowercase}
  
\fi
\IfFileExists{upquote.sty}{\usepackage{upquote}}{}
\IfFileExists{microtype.sty}{%
\usepackage{microtype}
\UseMicrotypeSet[protrusion]{basicmath} 
}{}
\ifxetex
  \usepackage[setpagesize=false, 
              unicode=false, 
              xetex]{hyperref}
\else
  \usepackage[unicode=true]{hyperref}
\fi
\usepackage[usenames,dvipsnames]{color}
\hypersetup{breaklinks=true,
            bookmarks=true,
            pdfauthor={Clay McLeod, University of Mississippi, \{clmcleod\}@go.olemiss.edu},
            pdftitle={A Framework for Distributed Deep Learning Layer Design in Python},
            colorlinks=true,
            citecolor=blue,
            urlcolor=blue,
            linkcolor=magenta,
            pdfborder={0 0 0}}
\usepackage{color}
\usepackage{fancyvrb}

\DefineVerbatimEnvironment{Highlighting}{Verbatim}{commandchars=\\\{\}}
\newenvironment{Shaded}{}{}
\newcommand{\KeywordTok}[1]{\textcolor[rgb]{0.00,0.44,0.13}{\textbf{{#1}}}}

\newcommand{\OtherTok}[1]{\textcolor[rgb]{0.00,0.44,0.13}{{#1}}}

\newcommand{\NormalTok}[1]{{#1}}
\usepackage{graphicx,grffile}
\makeatletter
\def\maxwidth{\ifdim\Gin@nat@width>\linewidth\linewidth\else\Gin@nat@width\fi}
\def\maxheight{\ifdim\Gin@nat@height>\textheight\textheight\else\Gin@nat@height\fi}
\makeatother
\setkeys{Gin}{width=\maxwidth,height=\maxheight,keepaspectratio}
\providecommand{\tightlist}{%
  \setlength{\itemsep}{0pt}\setlength{\parskip}{0pt}}
\title{A Framework for Distributed Deep Learning Layer Design in Python}
\author{Clay McLeod, University of Mississippi, \{clmcleod\}@go.olemiss.edu}
\date{October 2015}

\ifx\paragraph\undefined\else
\let\oldparagraph\paragraph
\renewcommand{\paragraph}[1]{\oldparagraph{#1}\mbox{}}
\fi
\ifx\subparagraph\undefined\else
\let\oldsubparagraph\subparagraph
\renewcommand{\subparagraph}[1]{\oldsubparagraph{#1}\mbox{}}
\fi

\begin{document}
\maketitle
\begin{abstract}
In this paper, a framework for testing Deep Neural Network (DNN) design
in Python is presented. First, big data, machine learning (ML), and
Artificial Neural Networks (ANNs) are discussed to familiarize the
reader with the importance of such a system. Next, the benefits and
detriments of implementing such a system in Python are presented.
Lastly, the specifics of the system are explained, and some experimental
results are presented to prove the effectiveness of the system.
\end{abstract}

\section{Introduction}\label{introduction}

In recent years, the amount of data that is produced annually has
increased dramatically {[}1{]}, {[}2{]}. The availability of this data
has prompted researchers to publish large amounts of literature that
study how we can make sense of this data, and, in turn, how can we
handle this data in a computationally efficient manner {[}3{]}, {[}4{]},
{[}5{]}, {[}6{]}. The best performing methods involve a statistical
technique called Machine Learning (ML), wherein we try to create a model
of the data by taking advantage of a machine's high processing power
{[}7{]}, {[}8{]}. For a complete overview of ML and the different
algorithms involved, please refer to {[}9{]}. One particularly good ML
algorithm for working with large amounts of data is a specific subset of
Artificial Neural Networks (ANNs) {[}10{]}, {[}11{]}, called Deep Neural
Networks (DNNs) {[}12{]}, {[}13{]}, {[}14{]}. DNNs work well with large
sets of data because of their peculiar quality of being a universal
approximator {[}15{]}, {[}16{]}, which means that theoretically
speaking, they can model any real function. This is extremely helpful in
the case of big data because DNNs can flush out relationships within
large amounts of data that humans would miss, due to our lower
computational capacity. Lately, Python has become a language of choice
for statisticians to model DNNs, partially because it abstracts away
difficult language semantics, allowing researchers to focus mainly on
the design of the algorithms rather than syntactical errors. Another key
factor in Python's success in the DNN research space is due to it's
portability and the large amount of scientific libraries that are
actively being developed, such as Numpy {[}17{]}, Scipy {[}18{]}, Pandas
{[}19{]}, and Theano {[}20{]}, {[}21{]}. This ecosystem allows for high
performance APIs with GPU acceleration to be run on a variety of
different operating systems. In this paper, we will examine the benefits
and detriments of using Python to model DNNs. Next, we will examine best
practices for implementing a scalable infrastructure within which adding
workers to the cluster is trivial. Lastly, some experimental results
will be presented to show the effectiveness of this approach.

\section{Machine Learning in Python}\label{machine-learning-in-python}

\subsection{Benefits}\label{benefits}

\subsubsection{Simplicity}\label{simplicity}

First, Python is a good language because it is an easily language to
understand and program with. Because the barrier to begin programming in
Python is so low, it attracts a large audience of people - especially
experts from other academic backgrounds who have little programming
experience but would like to take advantage of advanced computational
processes.

\subsubsection{Strong, Open Source
Community}\label{strong-open-source-community}

Partially because of the simplicity referenced earlier, a strong
community of academics has formed around the open source community. This
allows libraries to evolve and improve at a rapid pace that is not
reasonable when a small group of programmers is working on a closed
source project. Thus, many Python libraries utilize cutting edge
techniques and strong documentation pages. Many Python libraries also
use nonrestrictive licenses, which also prompts businesses to use them
and contribute back to the community.

\subsubsection{Mature scientific
libraries}\label{mature-scientific-libraries}

Furthermore, many experts in the scientific fields (wishing to reach the
largest possible audience) focus exclusively on developing libraries in
Python. This has lead to a mature, production tested ecosystem for
developing applications in Python (see ``Dependencies'').

\subsection{Detriments}\label{detriments}

\subsubsection{Global Interpreter Lock}\label{global-interpreter-lock}

One of the major roadblocks for implementing massively parallel systems
in Python in the existence of the Global Interpreter Lock (GIL)
{[}22{]}. The GIL is necessary in the default Python implementation
because Python's memory management system is not thread safe.
Concretely, this means that means that no two Python threads can access
a python object at the same time, which allows for concurrency, but not
parallelism {[}23{]}. This restriction greatly reduces Python's ability
to utilize multiple CPUs {[}24{]}. Furthermore, since the GIL has been
implemented in Python since its inception, many other subsystems within
Python's ecosystem have grown to rely on it's existence. Figure 1 shows
the some core Python scientific libraries and their relationship with
the GIL. Although several transpilers, such as IronPython and Jython,
have overcome Python's reliance on the GIL, these transpilers have not
been widely adopted due to their lack of full API support. No one has
successfully implemented a solution for the GIL, so the problem still
persists today.\footnote{Note that a Python interpreter is just a C
  level program. Whenever this C level program accesses a Python object,
  the GIL is locked for that process. However, different processes on a
  UNIX system have exclusive memory spaces, so each subprocess of Python
  has it's own GIL. This alleviates some of the problems introduced by
  the GIL, but objects still have serialized and deserialized for
  objects to communicate, which can be costly. Credit: email
  correspondence with Matthew Rocklin}

\begin{figure}[htbp]
\centering
\includegraphics{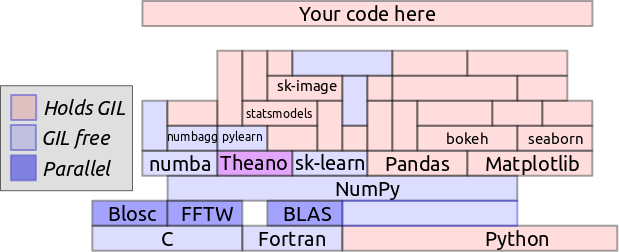}
\caption{Python scientific libraries and their reliance on the GIL
{[}23{]}}
\end{figure}

\section{Objectives}\label{objectives}

Before system design is discussed, the objectives of this paper and the
system should first be discussed. Many key goals were taken into
consideration during the design of this system, namely:

\begin{itemize}
\item
  \emph{Best DNN layer design}: Although some literature exists on the
  practical design of deep neural networks {[}25{]}, {[}26{]}, {[}27{]},
  designing deep neural networks remains a mix of following these
  guidelines and mastering the ``black art'' of designing deep neural
  networks. This system was designed to discover what empirical design
  rules can be discovered about designing deep neural networks by
  systemically evaluating performance of (1000-50000) DNNs on the same
  dataset while varying the hyper-parameters and layer design.
\item
  \emph{Simplicity}: When designing large systems such as this, one
  might lose sight of the forest for the trees. In this implementation,
  simplicity is the stressed wherever possible. This allows the user to
  focus solely on the design of the DNN rather than fix system errors.
\item
  \emph{Fullstack Python}: For this implementation, a fullstack Python
  solution will be presented. This is not necessarily for performance
  benefits so much as it is for completeness: this objective is
  important because the reader is presumably familiar with Python, but
  not necessarily familiar with any other programming languages.
  Therefore, to appeal to as wide of an audience as possible, only
  Python will be used in this system. Optimization of different parts of
  this system are left as an exercise to the reader.
\item
  \emph{Dataset}: Briefly, we are focusing on a very particular type of
  dataset. Namely, datasets under 1TB that contain numerical features
  and are trying to predict a certain class for the label. Generally
  speaking, this system could potentially work for other variants of
  data (\textgreater{} 1TB, regression problems). Due to the difficulty
  of addressing every type of problem that DNNs can solve exhaustively,
  those are left as an exercise to the reader. \emph{Note}: For datasets
  larger than 1TB, this system will probably note work well for two
  reasons. First, uploading of a CSV bigger than 1TB is extremely
  cumbersome through a web api, usually causing modern web browsers to
  crash. Second, the system design proposed assumes that your dataset
  will fit trivially in memory. If your dataset is larger than 1TB, the
  reader is encouraged to use a different system, such as Hadoop
  {[}5{]}.
\item
  \emph{Easy deployment}: Another key goal of this system design was to
  ensure ease/quickness of deployment across the system when testing
  rules and other code changes were applied to the code base. Since most
  researchers studying DNN design are not expert system administrators,
  this step is crucial to ensuring efficient research on DNNs.
\item
  \emph{GPU acceleration}: Although not true for every machine, some
  machines are equipped with graphic processing units (GPUs), which can
  be utilized to greatly decrease running time for training DNNs
  {[}28{]}. Therefore, this system ensures that each machine that has a
  GPU available will take advantage of this asset.
\end{itemize}

\section{Dependencies}\label{dependencies}

In the following sections, the \textbf{host machine} refers to a single
machine that coordinates the cluster of worker machines. This includes
distributed jobs through the message queue (see ``RabbitMQ''), the
database (see ``MongoDB''), and the HTTP reporting server (see
``Reporting Server''). A \textbf{worker machine} is a dispensable
machine that runs tasks as it receives it from the host machine. This
system has only been tested on Python 2.7, but porting to Python 3.X
should be trivial.

\subsection{Ecosystem}\label{ecosystem}

\subsubsection{Docker}\label{docker}

Docker is a lightweight Linux container system that allows you to
compile a specific Linux image once and deploy that image to multiple
destinations with ease {[}29{]}. Concretely, we can compile a Ubuntu
Linux image with the correct Python libraries installed then deploy our
image to each of our worker servers with ease. Docker runs on top of
most (if not all) major operating systems and takes up little overhead.
For more information on configuring Docker or its capabilities, visit
{[}30{]}.

\subsubsection{CoreOS}\label{coreos}

CoreOS is a lightweight, flexible operating system that runs containers
on top of it {[}31{]}. CoreOS has auto-clustering and job deployment
(called services) baked in. In addition, the CoreOS team has built a
painless distributed key-value store called \textbf{etcd} {[}32{]} that
comes with the operating system. Although many of these features are not
applicable to this particular paper, utilization of these features could
further improve upon the proposed implementation.

\subsubsection{RabbitMQ}\label{rabbitmq}

RabbitMQ is a robust message queue implementation that is easy,
portable, and supports a number of messaging protocols, including the
popular AMPQ protocol {[}33{]}. Message queues {[}34{]} are crucial when
building a distributed computing platform utilizing Python, because it
provides a centralized pool of tasks for each of your worker machines to
pull from. The decision to use this particular messaging queue to store
future parameters is based on the reputation of RabbitMQ being a
battle-tested, production ready product. However, other similar task
storing systems, such as Redis {[}35{]}, are worth mentioning.

\subsubsection{MongoDB}\label{mongodb}

MongoDB is a NoSQL database that offers easy storage and installation
{[}36{]}. This is one of the few components of the proposed system that
is chosen purely by preference instead of some performance based metric.
As previously stated, one of the major goals of this project is to keep
the solution as simple as possible. MongoDB helps achieve this goal
because of the vast amount of documentation available, the ease of
installation/management, and the availability of well-documented
libraries to interact with it. However, since MongoDB is only used for
results storage, any database you like could be used here. A
comprehensive comparison of databases and their trade offs can be found
in {[}37{]}.

\subsection{Reporting Server}\label{reporting-server}

The reporting server was built on a web platform, utilizing HTML5, CSS3,
and JavaScript. Notable libraries and their functions include:

\begin{itemize}
\tightlist
\item
  NGINX (\url{http://www.nginx.com/}): Serve static web pages and
  reverse proxy reporting server.
\item
  Bootstrap (\url{http://getbootstrap.com/}): CSS Styling library
  provided by Twitter.
\item
  jQuery (\url{http://jquery.com}): DOM manipulation and communication
  with backend REST server.
\item
  plot.ly (\url{http://plot.ly}): Plotting results, manipulating graphs,
  and exporting results.
\item
  Papa Parse (\url{http://papaparse.com/}): JS library for easily
  parsing CSV files.
\end{itemize}

\subsection{Python Libraries}\label{python-libraries}

\subsubsection{Flask}\label{flask}

Flask is a microframework for building web services uses Python
{[}38{]}. In this system, Flask acts as the middleman between the
website and the database by exposing a REST api {[}39{]} (see ``System
Design''). Although not suitable for larger project because of its
long-blocking operations, Flask was used in this system because of its
simplicity and the commitment to write as much of the system as possible
using fullstack Python.

\subsubsection{Celery}\label{celery}

Celery is an ``asynchronous task queue/job queue based on distributed
message passing'' {[}40{]}. Celery is crucial for the system presented
as it abstracts away all the intricacies of building a distributed
worker system, exposing an easy to use API for writing distributed
Python applications. Furthermore, there are several niceties built in,
such as compatibility with many major message queues/databases (namely
RabbitMQ, Redis, MongoDB) and web interfaces for monitoring performance.

\subsubsection{NumPy, SciPy, Pandas}\label{numpy-scipy-pandas}

NumPy {[}17{]}, SciPy {[}18{]}, and Pandas {[}19{]} are the foundational
scientific libraries upon which most other major Python libraries are
built. These libraries provide low level optimizations for scientific
computation across platforms. Furthermore, adoption of a common set of
libraries allows for easy interaction between different libraries.
Concretely, NumPy provides basic array functionality, SciPy provides a
suit of scientific functions and calculations (such as optimization and
statistical operations), and Pandas provides a common interface for
formatting and manipulating data. Of course, each of these does much
more than these functions, and the reader is encouraged to read their
respective documentations to gain a comprehensive list of the benefits
for using each.

\subsubsection{Theano}\label{theano}

Theano is a symbolic Python library that allows you to ``define,
optimize, and evaluate mathematical expressions with multi-dimensional
arrays efficiently'' {[}20{]}, {[}21{]}. Developed at the Université de
Montréal by leading DNN researchers, Theano plays a crucial role in the
DNN development stack. Theano allows easy GPU integration, provides
several key deep neural network functions (such as activation
functions), and, because of its symbolic natural, greatly simplifies the
math for the user by automatically computing complicated gradient
functions, as well as other complex mathematic equations.

\subsubsection{PyBrain}\label{pybrain}

PyBrain aims to offer a high performance neural network library without
the complicated entry barrier that most users experience when trying to
enter the field {[}41{]}. Developed independently at Technische
Universität München, PyBrain does \textbf{not} rely on Theano, and
therefore, does not take advantage of the GPU acceleration or optimized
mathematical functions Theano provides. However, it has been selected as
one of the DNN libraries because of its flexibility and ease of use.

\subsubsection{Keras}\label{keras}

Keras lives at the other end of the neural network library spectrum
{[}42{]}. Keras \textbf{is} built on top of Theano, and is closely
integrated with Theano under the hood. It takes full advantage of
Theano's optimizations wherever possible, and encourages its users to
use best practices and the latest DNN techniques. The barrier to learn
Keras and realize its full potential is slightly higher than that of
PyBrain. However, given its expressive and powerful nature, the tradeoff
is well worth it.

\section{System Design}\label{system-design}

For the purposes of this discussion, there are four atomic units of
functionality, all working asynchronously with each other. These units
of functionality include uploading the data (see ``Data Upload),
initializing tasks to run for that data (see''Task Management``),
executing the tasks (see''Task Execution``), and storing/visualizing the
results (see''Results Storage and Visualization``). Because these units
operate independently of each other, failure in the pipeline can occur
in one stage of the pipeline without affecting the performance of
another stage. Each stage''fails forward``, meaning that if a failure
occurs, the system merely presents the appropriate error and keeps
executing its next task rather than crashing. These are all principles
the reader is encouraged to follow when creating their own systems.

\subsection{Data Upload}\label{data-upload}

First, a static website was created for uploading data to the server to
be processed. This server used a standard Bootstrap template and exposed
an interface to upload a CSV file, since most datasets are easily
represented as this type of file. Next, the Papa Parse library was used
to parse the CSV file to a multidimensional JavaScript array. If Papa
Parse encountered an error in parsing the file (such as incorrect
format), it would through an error, which would be communicated to the
user and the process would be aborted. \emph{Note}: missing data was not
considered an error, due to the desired compatibility with sparse
datasets. If the dataset \emph{was} properly formatted, the user was
prompted to choose a ``label'' (what value of the data we would like to
predict), and Papa Parse would then hand off the data the jQuery, which
would attempt to upload the data to the Flask REST api. Upon success,
jQuery received a ``Session ID'' from the server, which was a unique
identifier that could be used to check on the progress/results of any
given session. At this point, the function of the web interface is
mostly done - if the user chooses to stay on the page, jQuery
continuously checks back with the server on the progress of that session
and updates a progress bar accordingly. When the current session
completes, the user is presented a link to review the results
visualization. Figure 2 shows the flowchart for the web interface.

\begin{figure}[htbp]
\centering
\includegraphics{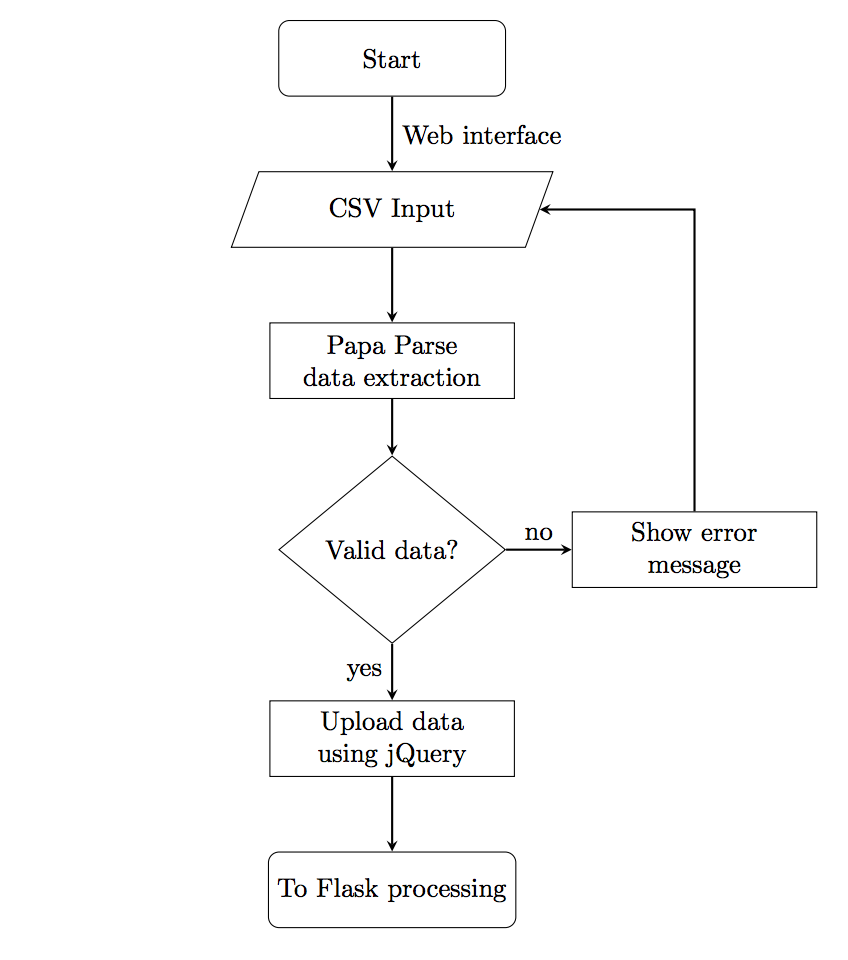}
\caption{Flowchart showing web interface process}
\end{figure}

Although we have discussed the function of the static website,
deployment of this website has not yet been discussed. This is where
Docker plays a crucial role in flexible deployment. First, a Docker
image was compiled from the base NGINX image. Concretely, Docker has
several precompiled containers available at {[}43{]} that you can base
your image off of. In this particular image, NGINX is already
preconfigured and running on top of the Ubuntu operating system. Simply
copy your static website files into the location specified on the NGINX
container page and you have a fully portable image ready to serve your
static website.

In this system, the Docker image serving our website is run on top of
the host machine for low latency communications with the database and
REST api. Port 80 of the host machine is mapped to the port that NGINX
is serving the website on inside of the container, meaning anyone who
visited the host machine's public IP address would get served the
website.

\subsection{Tasks Management}\label{tasks-management}

It has been hinted that the relationship between RabbitMQ and Celery is
an integral part of this system. As discussed earlier, setting up a
RabbitMQ server is very easy using Docker - simply run the default
RabbitMQ container provided at {[}43{]} with the web interface enabled.
In this system, the RabbitMQ container was run on the host machine for
low latency communication with the Flask server. For the Celery
instance, a separate Docker container was created on top of the default
Python 2.7 container {[}43{]} that handles the running of the
Flask/Celery Python microframework. Because all of these Docker
containers are run on the same machine and the ports are mapped to the
host machine, communication between Docker containers is seamless.

Once this JavaScript array has been passed off to the Flask server, the
user can preprocess the data as the like to prepare it for training.
This unit of functionality (as well as ``Task Execution'') is where most
of the ``black magic'' of designing DNNs comes into play. In this
system, a few simple experiments were designed to test different
hyper-parameters and layer designs across different datasets. Although
the specific of that design will be discussed here, the reader is
encouraged to come up with their own rules for designing their DNNs.

First, missing values were filled with zeroes (necessary for processing
in neural networks). Next, some best practices for working with DNNs
were implemented:

\begin{enumerate}
\def\labelenumi{\arabic{enumi}.}
\item
  All of the features (every value in the dataset except for the label)
  were scaled from {[}0..1{]} {[}25{]}.
\item
  Label data was split into categories and encoded using a technique
  called ``One Hot Encoding''. This creates a different output node in
  the DNN for each categorical answer and allows the DNN to output a
  probability that any given categorical answer is the correct one
  {[}25{]}.
\item
  The data was split into two distinct sets: 80\% training and 20\%
  testing. Holding out some data for testing a DNN helps to alleviate
  overfitting {[}44{]}.
\end{enumerate}

After the data has been preprocessed, it can be passed off to different
subsystems in the ``Task Execution'' unit for processing by Keras and
PyBrain. This is easily accomplished by creating two distinct celery
tasks for training a DNN with PyBrain and Keras respectively. Depending
on which library you would like to train with, you can submit jobs with
varying hyper-parameters and layers for processing in this stage of the
pipeline. Celery will serialize of this data into RabbitMQ and execute
these training tasks as worker machines become available. Figure 3 shows
the flowchart for the Task Management pipeline.

\begin{figure}[htbp]
\centering
\includegraphics{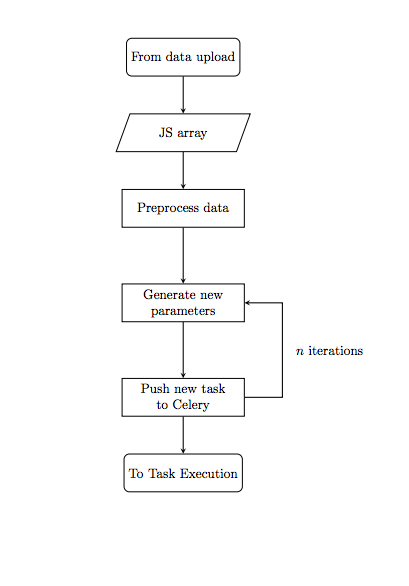}
\caption{Flowchart showing task management process}
\end{figure}

\subsection{Task Execution}\label{task-execution}

As discussed in ``Task Management'', this unit of operation simply
implements a training function for either Keras/PyBrain that trains the
DNN based on the parameters provided by Celery. This is a simple as
writing a single method and letting Celery run these training methods as
the workers become available. What is not trivial, however, is the rapid
deployment on new workers into the system.

A Docker container was built on top of the default Python 2.7 image
{[}43{]} that ran an instance of a Celery worker. In this way, anytime
you deploy this Docker container to a worker machine, another worker was
made available in the Celery worker cluster. The key to optimizing
efficiency is to take advantage of:

\begin{enumerate}
\def\labelenumi{\arabic{enumi}.}
\tightlist
\item
  Worker machines with GPU acceleration.
\item
  Worker machines with multiple cores.
\end{enumerate}

To take advantage of multiple CPU cores, simply set the
``THEANO\_FLAGS'' environment variable in the Docker as follows:

\begin{Shaded}
\begin{Highlighting}[]
\OtherTok{THEANO_FLAGS=}\NormalTok{device=}\KeywordTok{gpu}\NormalTok{,floatX=float32}
\end{Highlighting}
\end{Shaded}

You can specific this when you are deploying on a machine you know has a
GPU, or you can just always set the environment variable, as Theano
recognizing when the machine you are running is not configured with a
GPU.

One of the nice features about using a task scheduler like Celery is
that is has built in multi-process support. Meaning that you can simply
pass in a command line argument indicating the number of worker
processes you would like when starting a Celery worker is started, and
Celery will handle the logistics for you. For most use cases, a number
of workers equal to the number of cores on your worker machine is
recommended for optimal performance. Another alternative is to spawn
multiple Docker containers with a lower amount of workers, although this
approach is not quite as memory efficient due to the overhead of running
multiple Ubuntu instances. Figure 4 shows the flowchart for the Task
Execution process.

\begin{figure}[htbp]
\centering
\includegraphics{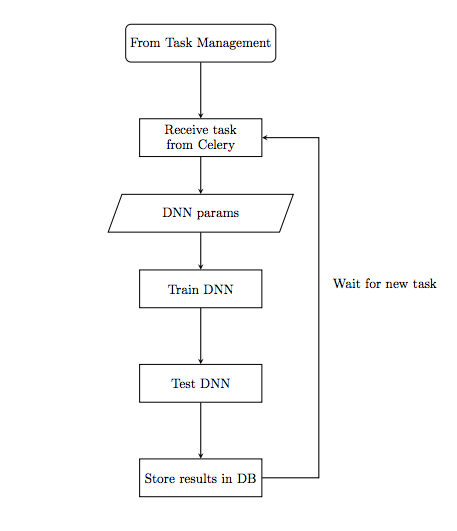}
\caption{Flowchart showing task execution process}
\end{figure}

\subsection{Results Storage and
Visualization}\label{results-storage-and-visualization}

After processing by the Task Execution subsystem, results are stored in
the MongoDB instance. A default MongoDB container {[}43{]} was
instantiated on the host machine, allowing for low latency communication
with the Flask server and a centralized datastore. A RESTful API was
used to access the MongoDB results from the client. Some information
stored in the database includes the session id, the training time, the
model accuracy, and the parameters used to train the model. Other
niceties of this setup include a web interface for monitoring both
Celery and RabbitMQ. Figures 5-7 show these reporting features.

\begin{figure}[htbp]
\centering
\includegraphics{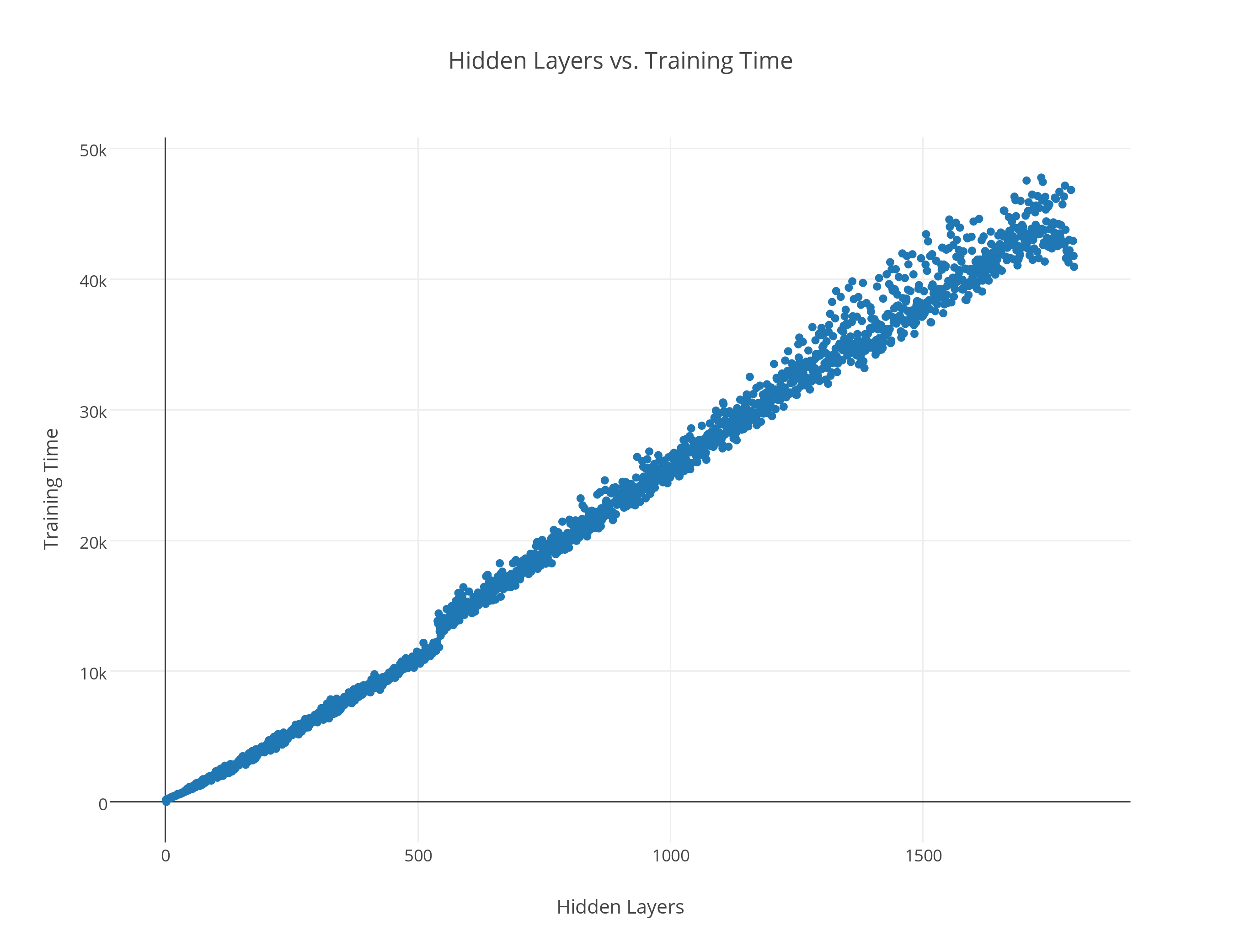}
\caption{Example of plot.ly showing training time vs.~hidden layers}
\end{figure}

\begin{figure}[htbp]
\centering
\includegraphics{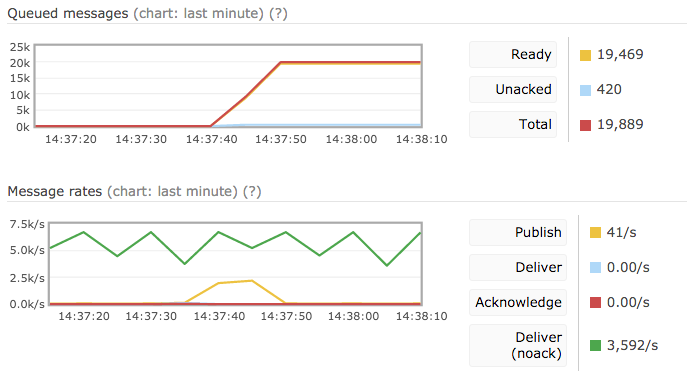}
\caption{RabbitMQ dashboard when uploading 20,000 jobs}
\end{figure}

\begin{figure}[htbp]
\centering
\includegraphics{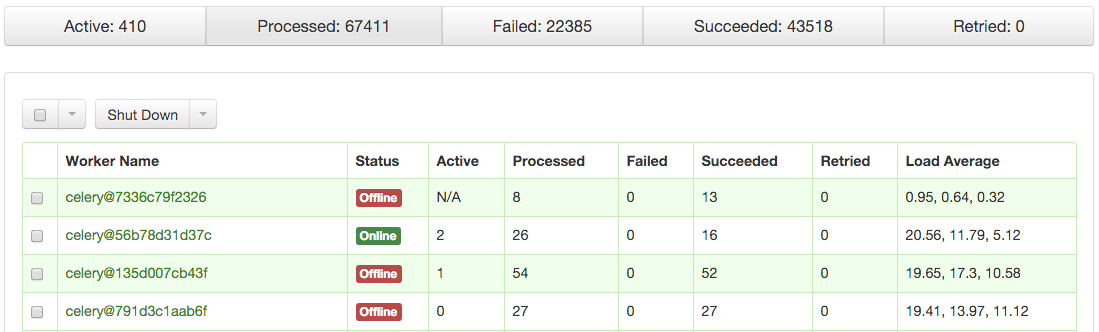}
\caption{Celery dashboard showing worker status}
\end{figure}

\section{Discussion}\label{discussion}

\subsection{Results}\label{results}

Discussion of results is not the main focus of this paper, and research
on what a systematic approach to studying DNNs is still ongoing.
However, a few preliminary observations stand out for those interested:

\begin{itemize}
\item
  Every DNN tested, on small datasets and large datasets, seem to reach
  a ``critical mass'' of information storage around 500-700 hidden
  layers, wherein the DNNs performance will flatline for any number of
  layers greater than the critical mass point. Quite probably, this is
  due to overfitting of the deep neural network to the training data and
  is not a significant finding, except to encourage designers to use as
  few layers as possible to avoid overfitting. \emph{Note}: this premise
  was not exhaustively tested against all kinds of layers and activation
  functions, but a significant number of combinations were tried.
\item
  Training of DNN increases roughly linearly with the amount of layers
  added to the network for a significant number of layer combinations
  tried (see Figure 5).
\item
  Granular control over parameters greatly increases performance in
  DNNs. For instance, testing several different combinations of
  activation functions will produce some interesting results.
\end{itemize}

\subsection{Improvements/Future Work}\label{improvementsfuture-work}

Several improvements are possible, notably:

\begin{itemize}
\item
  \emph{Asynchronous REST API}: This focus of this experiment was to
  build a full stack Python system to train DNNs. However, this is
  generally not optimal for asynchronous applications such as a REST
  API. The reader is encouraged to look for alternatives outside of the
  Python programming language, such as NodeJS {[}45{]} or golang
  {[}46{]}, to build their RESTful API, as Python's blocking nature
  presents major speed issues.
\item
  \emph{Granular web interface}: Rather than programming the rules for
  choosing training parameters directly, it is conceivable that you
  could build this functionality into the web user interface, allowing
  DNN designers not familiar with a specific library to set these
  training rules through a GUI.
\end{itemize}

\section{Conclusion}\label{conclusion}

To conclude, a distributed, fullstack Python implementation was
described in detail for examining performance results for different
parameters in DNN training. This is potentially useful because the
design of DNNs is more of an art than a science, so the more we can find
out about how different hyper-parameters/layers affect the accuracy of a
deep neural network, the better. By utilizing several deployment tools
(Docker {[}29{]}, CoreOS {[}31{]}), production grade distributed
solutions (RabbitMQ {[}33{]}, MongoDB {[}36{]}), and the Python
scientific ecosystem (NumPy {[}17{]}, SciPy {[}18{]}, Theano {[}20{]},
{[}21{]}, Keras {[}42{]}, PyBrain {[}41{]}), such a system is easily
implemented in a flexible way that prioritizes simplicity and
modularity. Some preliminary discoveries are discussed, and work to
build off of the current implementation is also presented.

\section*{References}\label{references}
\addcontentsline{toc}{section}{References}

\hyperdef{}{ref-big-data-1}{\label{ref-big-data-1}}
{[}1{]} J. Manyika, M. Chui, B. Brown, J. Bughin, R. Dobbs, C. Roxburgh,
and A. H. Byers, ``Big data: The next frontier for innovation,
competition, and productivity,'' 2011.

\hyperdef{}{ref-big-data-2}{\label{ref-big-data-2}}
{[}2{]} V. Mayer-Schönberger and K. Cukier, \emph{Big data: A revolution
that will transform how we live, work, and think}. Houghton Mifflin
Harcourt, 2013.

\hyperdef{}{ref-big-data-3}{\label{ref-big-data-3}}
{[}3{]} P. Russom and others, ``Big data analytics,'' \emph{TDWI Best
Practices Report, Fourth Quarter}, 2011.

\hyperdef{}{ref-big-data-4}{\label{ref-big-data-4}}
{[}4{]} P. Zikopoulos, C. Eaton, and others, \emph{Understanding big
data: Analytics for enterprise class hadoop and streaming data}.
McGraw-Hill Osborne Media, 2011.

\hyperdef{}{ref-hadoop}{\label{ref-hadoop}}
{[}5{]} K. Shvachko, H. Kuang, S. Radia, and R. Chansler, ``The hadoop
distributed file system,'' in \emph{Mass storage systems and
technologies (mSST), 2010 iEEE 26th symposium on}, 2010, pp. 1--10.

\hyperdef{}{ref-spark}{\label{ref-spark}}
{[}6{]} M. Zaharia, M. Chowdhury, M. J. Franklin, S. Shenker, and I.
Stoica, ``Spark: Cluster computing with working sets,'' in
\emph{Proceedings of the 2nd uSENIX conference on hot topics in cloud
computing}, 2010, vol. 10, p. 10.

\hyperdef{}{ref-ml-1}{\label{ref-ml-1}}
{[}7{]} D. E. Goldberg and J. H. Holland, ``Genetic algorithms and
machine learning,'' \emph{Machine learning}, vol. 3, no. 2, pp. 95--99,
1988.

\hyperdef{}{ref-ml-2}{\label{ref-ml-2}}
{[}8{]} J. G. Carbonell, R. S. Michalski, and T. M. Mitchell, ``An
overview of machine learning,'' in \emph{Machine learning}, Springer,
1983, pp. 3--23.

\hyperdef{}{ref-ml-book}{\label{ref-ml-book}}
{[}9{]} R. S. Michalski, J. G. Carbonell, and T. M. Mitchell,
\emph{Machine learning: An artificial intelligence approach}. Springer
Science \& Business Media, 2013.

\hyperdef{}{ref-nn-1}{\label{ref-nn-1}}
{[}10{]} S. Haykin and N. Network, ``A comprehensive foundation,''
\emph{Neural Networks}, vol. 2, no. 2004, 2004.

\hyperdef{}{ref-nn-2}{\label{ref-nn-2}}
{[}11{]} M. T. Hagan, H. B. Demuth, M. H. Beale, and others,
\emph{Neural network design}. Pws Pub. Boston, 1996.

\hyperdef{}{ref-dnn-1}{\label{ref-dnn-1}}
{[}12{]} J. Schmidhuber, ``Deep learning in neural networks: An
overview,'' \emph{CoRR}, vol. abs/1404.7828, 2014.

\hyperdef{}{ref-dnn-2}{\label{ref-dnn-2}}
{[}13{]} Y. Bengio, A. C. Courville, and P. Vincent, ``Unsupervised
feature learning and deep learning: A review and new perspectives,''
\emph{CoRR}, vol. abs/1206.5538, 2012.

\hyperdef{}{ref-dnn-book}{\label{ref-dnn-book}}
{[}14{]} Y. Bengio, I. J. Goodfellow, and A. Courville, ``Deep
learning,'' 2015.

\hyperdef{}{ref-universal-approximator-1}{\label{ref-universal-approximator-1}}
{[}15{]} K. Hornik, M. Stinchcombe, and H. White, ``Multilayer
feedforward networks are universal approximators,'' \emph{Neural
networks}, vol. 2, no. 5, pp. 359--366, 1989.

\hyperdef{}{ref-universal-approximator-2}{\label{ref-universal-approximator-2}}
{[}16{]} M. Leshno, V. Y. Lin, A. Pinkus, and S. Schocken, ``Multilayer
feedforward networks with a nonpolynomial activation function can
approximate any function,'' \emph{Neural networks}, vol. 6, no. 6, pp.
861--867, 1993.

\hyperdef{}{ref-numpy}{\label{ref-numpy}}
{[}17{]} N. team, ``NumPy GitHub repository,'' \emph{GitHub repository}.
GitHub, 2015.

\hyperdef{}{ref-scipy}{\label{ref-scipy}}
{[}18{]} E. Jones, T. Oliphant, P. Peterson, and others, ``SciPy: Open
source scientific tools for Python.'' 2001--2001-\/-.

\hyperdef{}{ref-pandas}{\label{ref-pandas}}
{[}19{]} W. McKinney, ``Data structures for statistical computing in
python,'' in \emph{Proceedings of the 9th python in science conference},
2010, pp. 51--56.

\hyperdef{}{ref-theano-1}{\label{ref-theano-1}}
{[}20{]} F. Bastien, P. Lamblin, R. Pascanu, J. Bergstra, I. J.
Goodfellow, A. Bergeron, N. Bouchard, and Y. Bengio, ``Theano: New
features and speed improvements.'' Deep Learning and Unsupervised
Feature Learning NIPS 2012 Workshop, 2012.

\hyperdef{}{ref-theano-2}{\label{ref-theano-2}}
{[}21{]} J. Bergstra, O. Breuleux, F. Bastien, P. Lamblin, R. Pascanu,
G. Desjardins, J. Turian, D. Warde-Farley, and Y. Bengio, ``Theano: A
CPU and GPU math expression compiler,'' in \emph{Proceedings of the
python for scientific computing conference (SciPy)}, 2010.

\hyperdef{}{ref-gil-doc}{\label{ref-gil-doc}}
{[}22{]} P. Foundation, ``Global interpreter lock documentation,'' 2015.
{[}Online{]}. Available:
\url{https://wiki.python.org/moin/GlobalInterpreterLock}. {[}Accessed:
23-Oct-2015{]}.

\hyperdef{}{ref-pydata-gil}{\label{ref-pydata-gil}}
{[}23{]} M. Rocklin, ``PyData and the gIL,'' 2015.

\hyperdef{}{ref-understanding-gil}{\label{ref-understanding-gil}}
{[}24{]} D. Beazley, ``Understanding the python gIL.'' 20-Feb-2010.

\hyperdef{}{ref-design-dnn-1}{\label{ref-design-dnn-1}}
{[}25{]} G. Hinton, ``A practical guide to training restricted boltzmann
machines,'' \emph{Momentum}, vol. 9, no. 1, p. 926, 2010.

\hyperdef{}{ref-design-dnn-2}{\label{ref-design-dnn-2}}
{[}26{]} Y. Bengio, ``Practical recommendations for gradient-based
training of deep architectures,'' in \emph{Neural networks: Tricks of
the trade}, Springer, 2012, pp. 437--478.

\hyperdef{}{ref-design-dnn-3}{\label{ref-design-dnn-3}}
{[}27{]} G. E. Hinton, N. Srivastava, A. Krizhevsky, I. Sutskever, and
R. R. Salakhutdinov, ``Improving neural networks by preventing
co-adaptation of feature detectors,'' \emph{arXiv preprint
arXiv:1207.0580}, 2012.

\hyperdef{}{ref-gpu}{\label{ref-gpu}}
{[}28{]} J. D. Owens, M. Houston, D. Luebke, S. Green, J. E. Stone, and
J. C. Phillips, ``GPU computing,'' \emph{Proceedings of the IEEE}, vol.
96, no. 5, pp. 879--899, 2008.

\hyperdef{}{ref-docker}{\label{ref-docker}}
{[}29{]} D. Merkel, ``Docker: Lightweight linux containers for
consistent development and deployment,'' \emph{Linux J.}, vol. 2014, no.
239, Mar. 2014.

\hyperdef{}{ref-docker-docs}{\label{ref-docker-docs}}
{[}30{]} D. Team, ``Docker documentation,'' 2015. {[}Online{]}.
Available: \url{http://docs.docker.com}. {[}Accessed: 24-Oct-2015{]}.

\hyperdef{}{ref-coreos}{\label{ref-coreos}}
{[}31{]} C. Team, ``CoreOS website,'' 2015. {[}Online{]}. Available:
\url{https://coreos.com/}. {[}Accessed: 24-Oct-2015{]}.

\hyperdef{}{ref-etcd}{\label{ref-etcd}}
{[}32{]} C. team, ``CoreOS GitHub repository,'' \emph{GitHub
repository}. \url{https://github.com/coreos/etcd}; GitHub, 2015.

\hyperdef{}{ref-rabbitmq}{\label{ref-rabbitmq}}
{[}33{]} R. Team, ``RabbitMQ website,'' 2015. {[}Online{]}. Available:
\url{https://www.rabbitmq.com/}. {[}Accessed: 24-Oct-2015{]}.

\hyperdef{}{ref-message-queue}{\label{ref-message-queue}}
{[}34{]} E. A. Brewer, F. T. Chong, L. T. Liu, S. D. Sharma, and J. D.
Kubiatowicz, ``Remote queues: Exposing message queues for optimization
and atomicity,'' in \emph{Proceedings of the seventh annual aCM
symposium on parallel algorithms and architectures}, 1995, pp. 42--53.

\hyperdef{}{ref-redis}{\label{ref-redis}}
{[}35{]} R. Team, ``Redis website,'' 2015. {[}Online{]}. Available:
\url{https://www.redis.io/}. {[}Accessed: 24-Oct-2015{]}.

\hyperdef{}{ref-mongodb}{\label{ref-mongodb}}
{[}36{]} M. Team, ``MongoDB website,'' 2015. {[}Online{]}. Available:
\url{https://www.mongodb.com/}. {[}Accessed: 24-Oct-2015{]}.

\hyperdef{}{ref-db-tradeoffs}{\label{ref-db-tradeoffs}}
{[}37{]} K. Kovacs, ``Cassandra vs mongoDB vs couchDB vs redis vs riak
vs hBase vs couchbase vs orientDB vs aerospike vs neo4j vs hypertable vs
elasticSearch vs accumulo vs voltDB vs scalaris vs rethinkDB
comparison,'' 2015. {[}Online{]}. Available:
\url{http://kkovacs.eu/cassandra-vs-mongodb-vs-couchdb-vs-redis}.
{[}Accessed: 24-Oct-2015{]}.

\hyperdef{}{ref-flask}{\label{ref-flask}}
{[}38{]} F. team, ``Flask website.'' 2015.

\hyperdef{}{ref-rest-api}{\label{ref-rest-api}}
{[}39{]} M. Masse, \emph{REST aPI design rulebook}. `` O'Reilly Media,
Inc.'', 2011.

\hyperdef{}{ref-celery}{\label{ref-celery}}
{[}40{]} C. team, ``Celery website.'' 2015.

\hyperdef{}{ref-pybrain}{\label{ref-pybrain}}
{[}41{]} T. Schaul, J. Bayer, D. Wierstra, Y. Sun, M. Felder, F. Sehnke,
T. Rückstieß, and J. Schmidhuber, ``PyBrain,'' \emph{Journal of Machine
Learning Research}, vol. 11, pp. 743--746, 2010.

\hyperdef{}{ref-keras}{\label{ref-keras}}
{[}42{]} F. et a. Chollet, ``Keras website,'' 2015. {[}Online{]}.
Available: \url{http://keras.io}. {[}Accessed: 24-Oct-2015{]}.

\hyperdef{}{ref-docker-hub}{\label{ref-docker-hub}}
{[}43{]} D. Team, ``Docker hub,'' 2015. {[}Online{]}. Available:
\url{http://hub.docker.com}. {[}Accessed: 25-Oct-2015{]}.

\hyperdef{}{ref-nn-overfitting}{\label{ref-nn-overfitting}}
{[}44{]} I. V. Tetko, D. J. Livingstone, and A. I. Luik, ``Neural
network studies. 1. comparison of overfitting and overtraining,''
\emph{Journal of chemical information and computer sciences}, vol. 35,
no. 5, pp. 826--833, 1995.

\hyperdef{}{ref-nodejs}{\label{ref-nodejs}}
{[}45{]} N. team, ``NodeJS website,'' 2015. {[}Online{]}. Available:
\url{http://nodejs.org/}. {[}Accessed: 25-Oct-2015{]}.

\hyperdef{}{ref-golang}{\label{ref-golang}}
{[}46{]} Google, ``Golang website,'' 2015. {[}Online{]}. Available:
\url{http://golang.org/}. {[}Accessed: 25-Oct-2015{]}.

\end{document}